%% file: main.tex
\documentclass[10pt,twocolumn,letterpaper]{article}

\PassOptionsToPackage{table}{xcolor}
\usepackage[pagenumbers]{cvpr} 
\usepackage{multirow}
\usepackage{graphicx}
\usepackage[normalem]{ulem}
\useunder{\uline}{\ul}{}


\definecolor{cvprblue}{rgb}{0.21,0.49,0.74}
\usepackage[pagebackref,breaklinks,colorlinks,allcolors=cvprblue]{hyperref}

\def\eg{\textit{e.g.}}

\def\paperID{2011} 
\def\confName{CVPR}
\def\confYear{2025}

\title{Visual-Oriented Fine-Grained Knowledge Editing for MultiModal Large Language Models}

\author{Zhen Zeng\\
Hefei University of Technology\\
Hefei, Anhui, China\\
{\tt\small zengzhen@mail.hfut.edu.cn}
\and
Leijiang Gu\\
Hefei University of Technology\\
Hefei, Anhui, China\\
{\tt\small 2024170839@mail.hfut.edu.cn}
\and
Xun Yang\\
University of Science and Technology of China\\
Hefei, Anhui, China\\
{\tt\small xyang21@ustc.edu.cn}
\and
Zhangling Duan\\
Hefei Comprehensive National Science Center\\
Hefei, Anhui, China\\
{\tt\small duanzl1024@ahu.edu.cn}
\and
Zenglin Shi\\
Hefei University of Technology\\
Hefei, Anhui, China\\
{\tt\small zenglin.shi@hfut.edu.cn}
\and
Meng Wang\\
Hefei University of Technology\\
Hefei, Anhui, China\\
{\tt\small eric.mengwang@gmail.com}
}

\begin{document}
\maketitle
\input{sec/0_abstract}    
\input{sec/1_intro}
\input{sec/2_related}

\input{sec/3_method}
\input{sec/4_benchmark}

\input{sec/5_experiment}
\input{sec/conclusion}
{
    \small
    \bibliographystyle{ieeenat_fullname}
    \bibliography{main}
}


\end{document}

%% file: sec/0_abstract.tex
\begin{abstract}
Knowledge editing aims to efficiently and cost-effectively correct inaccuracies and update outdated information. Recently, there has been growing interest in extending knowledge editing from Large Language Models (LLMs) to Multimodal Large Language Models (MLLMs), which integrate both textual and visual information, introducing additional editing complexities.
Existing multimodal knowledge editing works primarily focus on text-oriented, coarse-grained scenarios, failing to address the unique challenges posed by multimodal contexts. In this paper, we propose a visual-oriented, fine-grained multimodal knowledge editing task that targets precise editing in images with multiple interacting entities. We introduce the Fine-Grained Visual Knowledge Editing (FGVEdit) benchmark to evaluate this task.
Moreover, we propose a Multimodal Scope Classifier-based Knowledge Editor (MSCKE) framework. MSCKE leverages a multimodal scope classifier that integrates both visual and textual information to accurately identify and update knowledge related to specific entities within images. This approach ensures precise editing while preserving irrelevant information, overcoming the limitations of traditional text-only editing methods.
Extensive experiments on the FGVEdit benchmark demonstrate that MSCKE outperforms existing methods, showcasing its effectiveness in solving the complex challenges of multimodal knowledge editing.

\end{abstract}

%% file: sec/1_intro.tex
\section{Introduction}
\label{sec:intro}
Traditional knowledge editing primarily strives to update and correct the knowledge in Large Language Models (LLMs) to ensure their accuracy and reliability ~\cite{zhang2024comprehensive,yao2023editing,sinitsin2019editable}. Many works, \eg, \cite{serac, ike, madaan2022memory, mend, de2021editing,zhu2021modifying}, have proposed various knowledge editing methods for LLMs, which have shown strong performance in refining model knowledge. Recently, there has been a growing interest in extending knowledge editing to Multimodal Large Language Models~(MLLMs)~\cite{mmedit,yin2023survey}. MLLMs enhance the capabilities of traditional LLMs by incorporating multiple modalities, such as text and images, allowing them to process and generate content that integrates both textual and visual information~\cite{alayrac2022flamingo,minigpt,blip2,liu2024visual,yang2023mm,tsimpoukelli2021multimodal}. In MLLMs, errors can arise not only from the language module but also from the visual components or the interactions between modalities. As a result, knowledge editing in MLLMs presents unique challenges that go beyond the complexities found in LLMs.
\begin{figure}[t]
  \centering

   \includegraphics[width=0.95\linewidth]{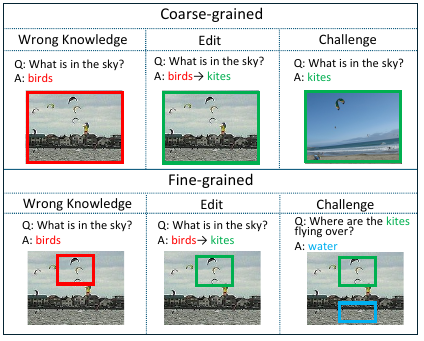}

   \caption{Comparison of fine-grained and coarse-grained knowledge editing in multimodal models. Coarse-grained knowledge editing treats the entire image as a single entity, allowing challenges to be addressed through simple text replacements. In contrast, fine-grained knowledge editing presents challenges that require editing methods to target specific entities within the image.
}
   \label{fig:intro}
\end{figure}

To advance the field, some pioneered works \cite{mmedit,kebench,mcmke,mike} have introduced multiple multimodal knowledge editing benchmarks. However, these benchmarks still primarily focus on text-oriented knowledge editing involving coarse-grained visual understanding, only considering single entities within images. In such settings, editing methods do not require access to the visual module; simply editing the language model suffices~(\eg, by replacing the word "bird" with "kite"). As a result, knowledge editing methods for LLMs tend to perform well within these settings. In contrast, this work addresses the more complex task of editing knowledge associated with multiple interacting entities within a single image. To this end, we introduce a new FGVEdit benchmark, designed for visual-oriented fine-grained knowledge editing. As illustrated in Fig.~\ref{fig:intro}, our problem setting requires editing methods to accurately identify and localize target entities within images, updating the knowledge tied to these entities (e.g., the kites and their associated relationships). In this context, traditional knowledge editing methods for LLMs are less effective.

In this work, we introduce the Multimodal Scope Classifier-based Knowledge Editor (MSCKE) framework, specifically designed for visual-oriented, fine-grained knowledge editing. Building upon the text-based editing method SERAC~\cite{serac}, we make a key modification by replacing the text-only scope classifier with a multimodal scope classifier. Our proposed classifier combines both visual and textual modalities to accurately identify fine-grained target entities and extract relevant knowledge from images, enabling the capture of subtle entity similarities. This multimodal approach ensures that edited knowledge is applied only when the new input is meaningfully related to the target entity being edited. Such integration is crucial, as relying solely on textual semantics may fail to capture the nuanced relationships between entities in the same image. Our contributions are summarized as follows:
\begin{itemize}
\item We introduce a new visual-oriented fine-grained knowledge editing task tailored for multimodal settings, emphasizing the unique challenges and characteristics of multimodal knowledge editing as distinct from traditional text-based approaches.
\item We present MSCKE, a new framework specifically designed for multimodal knowledge editing in MLLMs. It enables precise, fine-grained knowledge modifications by leveraging a multimodal scope classifier that integrates both visual and textual information.
\item We introduce the FGVEdit benchmark, featuring images with multiple entities. This benchmark is designed to evaluate the visual-oriented fine-grained editing capabilities of multimodal knowledge editing methods, pushing the limits of editing accuracy and relevance.
\end{itemize}
Extensive experiments on the FGVEdit benchmark demonstrate that our method outperforms existing approaches, demonstrating the effectiveness of the MSCKE framework in addressing the unique challenges of multimodal knowledge editing.

%% file: sec/2_related.tex
\section{Related Works}
\label{sec:related}

\subsection{Knowledge Editing for LLMs}
The existing knowledge editing methods for LLMs can be categorized into two main paradigms: parameter-preserving methods and parameter-modifying methods~\cite{lazaridou2021mind,touvron2023llama,achiam2023gpt,brown2020language}.

\noindent\textbf{Parameter-preserving methods}.
This approach modifies the output during inference of LLMs by utilizing additional components, while keeping the original model parameters unchanged. These additional components can include memory that stores edited examples and supplementary models that process the output. Representative methods include SERAC~\cite{serac}, IKE~\cite{ike}, and MemPrompt~\cite{madaan2022memory}. Additionally, T-Patcher~\cite{huangtransformer} and CaliNET~\cite{dong2022calibrating} incorporate extra trainable parameters into LLMs.

\noindent\textbf{Parameter-modifying methods}.
This approach involves updating certain parameters within MLLMs to modify the model. One intuitive strategy is to identify and modify the parameters corresponding to specific knowledge, as demonstrated by KN~\cite{dai2022knowledge}, ROME~\cite{meng2022locating}, and MEMIT~\cite{mengmass}. Another modification method employs meta-learning to train a hypernetwork that updates the parameters of LLMs, exemplified by MEND~\cite{mend} and KE~\cite{de2021editing}. 

Building upon the parameter-preserving text editing methods SERAC~\cite{serac}, we introduce the Multimodal Scope Classifier-based Knowledge Editor (MSCKE) framework by replacing the text-only scope classifier with a multimodal scope classifier. MSCKE establishes a new framework to extend knowledge editing from LLMs to MLLMs. 

\subsection{Knowledge Editing for MLLMs}
MLLMs are capable of processing information from different modalities, such as text and images \cite{clip, alayrac2022flamingo, yin2023survey}. Typically, they consist of a visual encoder, a language model, and an alignment module that integrates the visual and textual spaces. When performing knowledge editing on MLLMs, it is crucial to account for various types of errors, including visual comprehension errors, textual knowledge errors, and multimodal coordination errors.

MMEdit \cite{mmedit} is the first benchmark for multimodal knowledge editing, designed to evaluate the effectiveness of editing methods in modifying visual knowledge. KEBench \cite{kebench} introduces a generality metric to assess how well models leverage edited knowledge in relation to associated content. MIKE \cite{mike} and MC-MKE \cite{mcmke} propose fine-grained multimodal knowledge editing benchmarks at different levels. Specifically, MIKE \cite{mike} updates entity knowledge from coarse-grained to fine-grained (e.g., changing "old man" to "Trump"), while MC-MKE \cite{mcmke} differentiates between various types of errors.
\begin{figure*}[ht!]
  \centering
   \includegraphics[]{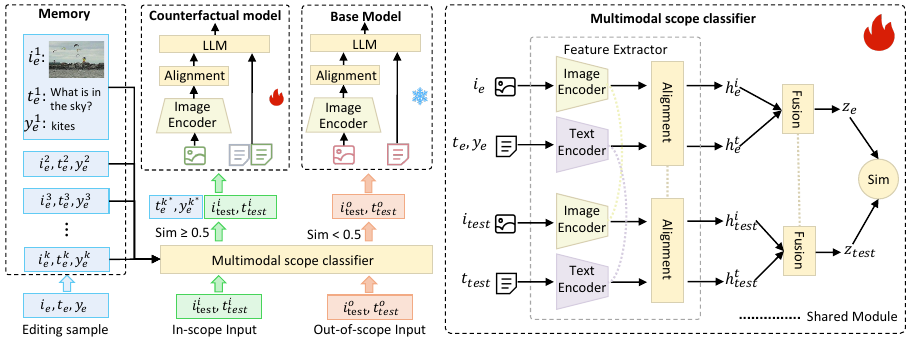}

   \caption{Architecture of the MSCKE method, illustrating the multimodal scope classifier, editing memory, base model with frozen parameters, counterfactual model with trainable parameters. During the editing phase, the editing samples are stored in the editing memory. In the inference phase, the multimodal classifier is employed to compute the similarity between the input and the editing samples in memory. Inputs with a similarity score less than 0.5 are classified as out-of-scope inputs and are processed by the base model; conversely, inputs are handled by the counterfactual model.
}
   \label{fig:method}
\end{figure*}

However, the existing works treat a single image as a single entity for editing, primarily focusing on textual knowledge. As a result, knowledge editing methods for LLMs tend to perform well within these settings. In contrast, this work focuses on the complex interactions between multiple entities within a single image and proposes a new FGVEdit benchmark for fine-grained visual knowledge editing. Under our problem setting, traditional knowledge editing methods for LLMs do not perform as effectively, prompting us to introduce a new MSCKE framework with a multimodal scope classifier.

%% file: sec/3_method.tex
\section{Fine-Grained Visual Knowledge Editing}

\subsection{Problem Formulation}
Fine-grained visual knowledge editing~(FGVEdit) task focuses on updating specific and detailed knowledge within MLLMs that process both visual and textual information. Unlike other multimodal knowledge editing tasks, which treat an image as an entity and perform coarse-grained edits that may overlook complex visual details, FGVEdit aims to modify knowledge related to specific entities and relationships involving multiple interactive elements within complex images.

Formally, let $ f_{\theta} $ denote a pre-trained MLLM with parameters $ \theta $ that can generate outputs based on combined image-text inputs. An example of the model's editing is represented as $ (i_e, t_e, y_e) $, where $ i_e $ denotes an image containing multiple entities, $ t_e $ is the textual prompt, and $ y_e $ represents the expected output for the input $ (i_e, t_e) $. The model's original output is given by $ y_o = f_{\theta}(i_e, t_e) $ and $y_o \neq y_e$.

The objective of the FGVEdit task is to update the model's knowledge so that it accurately reflects $ y_e $ when queried for specific fine-grained details, without affecting its performance on unrelated content. We define a knowledge editing operation $ \mathcal{E} $ to update the model parameters as $\theta_e = \mathcal{E}(\theta, i_e, t_e, y_e)$.
After applying $\mathcal{E}$, the updated model $ f_{\theta_e} $ should be capable of generating the corrected output for the specific input as $ y_e = f_{\theta_e}(i_e, t_e) $.

\noindent\textbf{Fine-grained visual editing scope.} For the fine-grained knowledge defined by the editing example $ (i_e, t_e, y_e) $, there exists a fine-grained editing scope $ S(i_e, t_e, y_e) $. This editing scope is determined by both visual and textual information. For in-scope inputs on the same image, represented as $ (i_{e}, t_{in}) \in S(i_e, t_e, y_e) $, the original model's output is $ y_{in} = f_{\theta}(i_e, t_{in}) $, and the edited model is required to produce a correction as $ y\prime_{in} = f_{\theta_e}(i_e, t_{in}) $. For out-of-scope inputs on the same image, represented as $ (i_e, t_{out}) \notin S(i_e, t_e, y_e) $, the original model's output is $ y_{out} = f_{\theta}(i_e, t_{out}) $, and the edited model maintains the original output as $ y_{out} = f_{\theta_e}(i_e, t_{out}) $.

\subsection{MSCKE Framework}
Building upon SERAC~\cite{serac}, we propose a Multimodal Scope Classifier-based Knowledge Editor~(MSCKE) framework, as illustrated in Fig.~\ref{fig:method}. Structurally, our MSCKE framework comprises the following four major components: the \textbf{Multimodal Edit Memory}, the \textbf{Multimodal Scope Classifier}, the \textbf{Base Multimodal Model} ($f_{\text{base}}$), and the \textbf{Counterfactual Multimodal Model} ($f_{\text{cfr}}$).

During the editing process, MSCKE does not modify the base model parameters. Instead, it stores the editing examples in the MultiModal Edit Memory. The MultiModal Edit Memory is a model-agnostic component designed to accept and store the knowledge that needs to be updated, represented as edit examples. These examples specify the precise modifications or corrections to the model’s knowledge. By maintaining these edit examples separately, MSCKE can reference them when necessary without altering the base model directly.

When receiving a new input, the MultiModal Scope Classifier evaluates the relevance of the input to the stored editing examples based on visual and textual information. It then decides whether to invoke the edited information or rely on the base model's response. This ensures that edits are applied appropriately to relevant inputs, while unrelated knowledge remains stable. The MultiModal Scope Classifier is defined as $f_{\text{cls}}(i_{\text{test}}, t_{\text{test}}, i_e, t_e, y_e)$.

The index $k^*$ of the most similar editing example and its corresponding similarity $\rho$ are determined as follows:
\begin{align}
k^* &= \mathop{\arg\min}\limits_k f_{\text{cls}}(i_{\text{test}}, t_{\text{test}}, i_e^k, t_e^k, y_e^k), \\
\rho &= f_{\text{cls}}(i_{\text{test}}, t_{\text{test}}, i_e^{k^*}, t_e^{k^*}, y_e^{k^*}),
\end{align}
where $\rho \in [0,1]$. Inputs with similarity less than $0.5$ are considered out-of-scope samples, while those with similarity equal to or greater than $0.5$ are regarded as in-scope samples. The final output of the model is:
\begin{align}
y_{\text{test}} = 
\begin{cases}
f_{\text{base}}(i_{\text{test}}, t_{\text{test}}), & \rho < 0.5, \\
f_{\text{cfr}}(t_e, y_e, i_{\text{test}}, t_{\text{test}}), & \rho \geq 0.5,
\end{cases}
\end{align}
where $f_{\text{base}}$ denotes the base model, and $f_{\text{cfr}}$ denotes the counterfactual model. 

The base multimodal model is a well-trained multimodal large language model (e.g., MiniGPT-4), designed to process inputs that are unrelated to the editing examples in the multimodal memory. By freezing the parameters of the base multimodal model, we ensure that its general knowledge remains stable and unaffected by the editing examples. In contrast, the counterfactual multimodal model responds to inputs related to the editing examples. Specifically, the counterfactual multimodal model accepts the input $(i_{\text{test}}, t_{\text{test}})$ while incorporating the corresponding editing examples from the multimodal editing memory. Guided by the prompts of the editing examples in memory, the counterfactual multimodal model generates outputs that reflect the necessary knowledge updates or corrections. The counterfactual multimodal model can either be a trainable model that shares the same output space as the base multimodal model or employ alternative editing methods that accurately reflect the new facts.

\subsection{Multimodal Scope Classifier}
Unlike the text editing scope classifier in SERAC, we have developed a new multimodal scope classifier within the MSCKE framework, designed to integrate both visual and textual modalities. This allows the classifier to capture fine-grained similarities between input and editing examples. Such integration is essential, as relying solely on textual semantics may fail to capture true similarity when $t_e$ and $t_{\text{test}}$ refer to different entities related to $i_e$ and $i_{\text{test}}$. By combining image and text information, our multimodal scope classifier enables a more comprehensive evaluation of the relevance between the query and editing examples. For both images and texts, we first map them into a unified feature space:
\begin{align}
h^i_e = A_i(E_i(i_e)),&\quad
h^t_e = A_t(E_t(t_e, y_e)),
\\
h^i_{\text{test}} = A_i(E_i(i_{\text{test}})),&\quad
h^t_{\text{test}} = A_t(E_t(t_{\text{test}})),
\end{align}
in which $E_i(\cdot)$ and $E_t(\cdot)$ are encoders for images and texts, respectively, which extract high-level features from the inputs. The Alignment modules $A_i(\cdot)$ and $A_t(\cdot)$ project the encoded features into a shared multimodal space, ensuring that the image and text representations are compatible for fusion. To simplify the process, we utilize a pre-trained CLIP~\cite{clip} model for fine-tuning, which is capable of extracting aligned features from both images and text. We then compute the image-text fused features:
\begin{align}
z_e &= \text{Fusion}(h^i_e, h^t_e),
\\
z_{\text{test}} &= \text{Fusion}(h^i_{\text{test}}, h^t_{\text{test}}),
\end{align}
in which the $\text{Fusion}(\cdot, \cdot)$ function combines the visual and textual features to capture the cross-modal interactions. To simplify the training process, we employ the dot-product attention mechanism to extract the portions of the image that are relevant to the text. Finally, the similarity is calculated from the fused features:
\begin{align}
\rho = \text{Sim}(z_e, z_{\text{test}}),
\end{align}
where $\text{Sim}(\cdot, \cdot)$ is a similarity function, such as cosine similarity or a learned distance metric, which quantifies the relevance between the query and the editing example in the fused feature space. We would highlight that our multimodal scope classifier can accept text-only input by setting the image input to null, ensuring compatibility with text editing tasks.

\noindent\textbf{Training}.
The multimodal scope classifier is treated as a binary classifier, designed to determine whether a given query is related to the editing examples stored in memory. Thus, we train the classifier with a binary cross-entropy loss on a multimodal dataset, denoted as $D_e = \{(i^{k}_e, t^{k}_e, y^{k}_e)\}$. For each editing sample, the dataset provides both in-scope inputs $(i^{k}_{in}, t^{k}_{in})$ and out-of-scope inputs $(i^{k}_{in}, t^{k}_{in})$. The loss function is defined as:
\begin{align}
\rho^{k}_{\text{in}} &= f_{\text{cls}}(i^{k}_{\text{in}}, t^{k}_{\text{in}}, i^{k}_e, t^{k}_e, y^{k}_e),\\
\rho^{k}_{\text{out}} &= f_{\text{cls}}(i^{k}_{\text{out}}, t^{k}_{\text{out}}, i^{k}_e, t^{k}_e, y^{k}_e),
\end{align}
\begin{equation}
\mathcal{L}_{cls} = -\frac{1}{N} \sum_{k=1}^{N} [ \log(\rho^{k}_{\text{in}}) + \log(1 - \rho^{k}_{\text{out}})],
\end{equation}
where $N$ is the total number of training samples.


%% file: sec/4_benchmark.tex
\begin{figure*}[ht]
  \centering
   \includegraphics[]{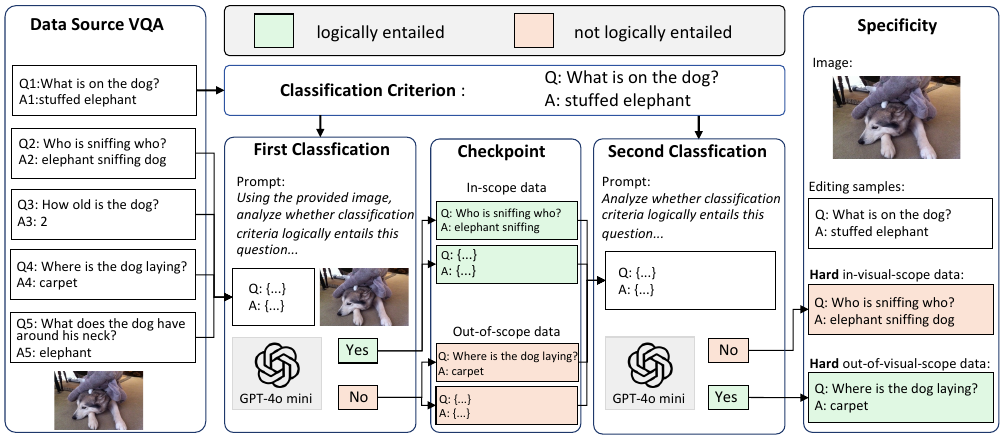}

   \caption{The construction pipeline of specificity dataset. For all questions related to a given image, the first question is selected as the classification criterion. Subsequent questions are assessed using ChatGPT. This classification process occurs in two stages: the first stage categorizes the data into in-scope and out-of-scope based on the image, while the second stage employs a text-based classification to filter out difficult data. Difficult data refers to instances where the category cannot be determined solely through textual analysis.
}
   \label{fig:bench}
\end{figure*}

\section{FGVEdit Benchmark}

\subsection{Evaluation Metrics}

\textbf{Specificity.}
We introduce a new metric, Specificity, specifically designed to evaluate fine-grained visual editing performance. Previous multimodal knowledge editing metrics treat an image as single entity and fail to account for the multiple entities present in the same image. These metrics primarily assess the editing method’s impact on the language model of MLLMs, rather than the nuanced interactions within the image itself. In contrast, Specificity, denoted as $M_{\text{specifity}}$, is capable of evaluating multiple entities and their interactions within the same image. To capture the different relationships between entities in an image, $M_{\text{specifity}}$ is computed by two components: $M^{\text{v}}_{\text{in}}$ and $M^{\text{v}}_{\text{out}}$. $M^{\text{v}}_{\text{in}}$ evaluates the model's ability to respond correctly to questions within the visual scope of the editing entity, while $M^{\text{v}}_{\text{out}}$ assesses the model's performance on questions outside the visual scope of the editing entity. These components are defined as follows:

\begin{gather}
 M_{specifity}=\frac{1}{2} [M_{in}^v+M_{out}^v]\\
 M_{in}^{v} =\mathbb{E} _{(t_{in},y\prime_{in})\sim D_{in}^v }[\mathbb{I}_{f_{\theta_{e}}(i_{e},t_{in})=y\prime_{in}}] \\
M_{out}^{v} =\mathbb{E} _{(t_{out},y_{out})\sim D_{out}^v }[\mathbb{I}_{f_{\theta_{e}}(i_{e},t_{out})=f_{\theta}(i_{e},t_{out})}] 
 \end{gather}  
where $D_{in}^v$ refers to dataset of in-visual-scope data, $D_{out}^v$ represents dataset of out-of- visual-scope data.

\noindent\textbf{Reliability.}
The reliability of the model refers to its ability to change its prediction from the original output $y_o$ to the desired output $y_e$ when provided with an image and text input. This is a key objective in knowledge editing. We measure reliability as follows:
\begin{align}
M_{rel}=\mathbb{E} _{(i_{e},t_{e},y_{e})\sim D_{edit} }[\mathbb{I}_{f_{\theta_{e}}(i_{e},t_{e})=y_{e}}] 
\end{align}
where $D_{edit} $ denotes to the dataset of Reliability.

\noindent\textbf{Locality.}
Locality measures the model’s stability, ensuring that it produces consistent outputs for data outside the editing scope before and after knowledge editing. We measure locality as follows:
\begin{align}
 M_{loc} =\mathbb{E}_{(t_{l},y_{l})\sim D_{loc} }[\mathbb{I}_{f_{\theta_{e}}(t_{l})=f_{\theta}(t_{l})}]                               
\end{align}
where $t_l$ represents question out of editing scope, $y_l$ is the corresponding label, $D_{loc} $ refers to dataset of Locality.

\noindent\textbf{Generality.}
In addition to accurately changing the prediction from the original $y_o$ to the desired $y_e$, the edited multimodal model should also adapt its predictions within the editing scope, such as for rephrased questions. We define generality as follows:
\begin{align}
M_{gen}  =\mathbb{E} _{(t_{r})\sim D_{gen} }
[\mathbb{I}_{f_{\theta_{e}}(i_{e},t_{r})=f_{\theta_{e}}(i_{e},t_{e})}]
\end{align}
where $t_r$ represents the rephrased question. $D_{gen} $ refers to dataset of Generality.

\subsection{Dataset Construction}
We select Visual Question Answering (VQA) as the evaluation task and use VQAv2 \cite{VQAv2} as the foundation for constructing our FGVEdit dataset. VQAv2 was originally designed to assess a model's ability to answer open-ended questions about images. It contains over 1 million question-answer pairs, each associated with an image, spanning a broad array of topics and requiring diverse reasoning capabilities.

\noindent\textbf{Specificity dataset construction.}
We select the first question of each image as the classification criterion. 
Additionally, it also serves as the editing sample for the calculation of reliability.
To categorize the remaining questions, we manually construct two prompts and apply them twice. First, using the classification criterion and the corresponding image, we request GPT-4o-mini with a prompt "\textit{Using the provided image, analyze whether classification criterion logically entails this question, specifically, whether a change in Answer 1 would impact Answer 2. Please make a simple judgment (Yes, No, Maybe), and your response should not contain any other characters.}", to classify the remaining questions, assessing whether the criterion logically entails the other questions. The results of this first classification yield two lists: In-scope samples, representing questions logically entailed by the first question, and Out-of-scope samples, representing questions that are not logically entailed by it. Next, based solely on the classification criterion, we request GPT-4o-mini again with a prompt "\textit{analyze whether classification criterion logically entails this question, specifically, whether a change in Answer 1 would impact Answer 2. Please make a simple judgment (Yes, No, Maybe), and your response should not contain any other characters.}", to classify the questions in In-scope samples and Out-of-scope samples, generating the final results: "hard in-visual-scope" for questions within the visual context of the image and "hard out-of-visual-scope" for questions outside the visual context.

\noindent\textbf{Locality and Generality dataset construction.}
For the Locality dataset, we follow the approach in MMEdit \cite{mmedit} and use the NQ dataset \cite{NQdataset}, a benchmark for model stability, which includes question-answer pairs that fall outside the editing scope. For the Generality dataset, we request GPT-4o-mini with a prompt like "\textit{Please rewrite the following question differently. Do not include the original question}" and assess whether the model can accurately respond to the generated questions within the editing scope.

\subsection{Data Statistics}
For each sample, we generate instances of specificity, locality, and generality data to ensure dataset completeness. Due to the limited availability of specificity data that includes both hard in-visual-scope and hard out-of-visual-scope data, we retain specificity data that contains either one of these categories. This approach maintains the task's challenge while ensuring sufficient data coverage. Non-hard data is used as a substitute for the other category when necessary. The final dataset consists of 11,112 samples, which are then split into training and testing sets in a 3:1 ratio.

%% file: sec/5_experiment.tex
\begin{table*}[t]
\renewcommand{\arraystretch}{1.05}
\centering
\resizebox{0.9\textwidth}{!}{%

\begin{tabular}{lcccccccc}
\hline
     & \multicolumn{4}{c}{BLIP-2 OPT}                                    & \multicolumn{4}{c}{MiniGPT-4}                                     \\
\cmidrule(lr){2-5} \cmidrule(lr){6-9}
    & Reliability    & Locality     & Generality   & Specificity  & Reliability    & Locality     & Generality   & Specificity  \\ \hline
FT-LLM    & \textbf{100.0} & 76.93          & \textbf{99.96} & 24.21          & 93.39          & 86.26          & 93.35          & 35.02          \\
FT-Visual & 99.68          & \textbf{100.0} & {\ul 99.21}    & 16.56          & 93.39          & \textbf{100.0} & 91.87          & 30.53          \\
IKE       & {\ul 99.89}    & 48.49          & 98.02          & 20.07          & \textbf{100.0} & 52.45          & \textbf{98.88} & 25.26          \\ \hline
SERAC     & 93.08          & {\ul 99.90}    & 96.83          & 31.92          & {\ul 99.50}    & \textbf{100.0} & 92.90          & 37.85          \\
\rowcolor{gray!20}
MSCKE      & 99.13          & \textbf{100.0} & 98.56          & 61.60          & {\ul 99.50}    & \textbf{100.0} & 93.00          & 57.20          \\ \hline
MEND      & 97.00          & 98.60          & 96.40          & {\ul 65.85}    & 94.85          & {\ul 98.58}    & 94.82          & {\ul 67.39}    \\
\rowcolor{gray!20}
MSCKE-MEND & 97.40          & \textbf{100.0} & 96.50          & \textbf{68.38} & 97.05          & \textbf{100.0} & {\ul 96.70}    & \textbf{71.98} \\ \hline
\end{tabular}%
}
\caption{Performance comparison of various knowledge editing methods on BLIP-2 OPT and MiniGPT-4. Results are presented as percentages, with the best results in \textbf{bold} and the second-best results \underline{underlined}.}
\label{tab:results}
\end{table*}

\section{Experiments and Results}
\subsection{Experimental Setup}
\noindent\textbf{Base MLLMs.}
We utilize BLIP-2 OPT~\cite{blip2} and MiniGPT-4~\cite{minigpt} as the base MLLMs for Editing.
\textbf{BLIP-2 OPT} is a vision-language model that combines a pre-trained Vision Transformer (ViT) with a lightweight Querying Transformer (Q-Former). It uses a two-stage pre-training strategy: the first stage focuses on learning vision-language representations, while the second emphasizes generative learning, allowing for efficient zero-shot image-to-text generation. 
\textbf{MiniGPT-4} integrates a pre-trained ViT with a Q-Former to effectively align visual features with the Vicuna LLM. This architecture enables it to process and generate text based on visual inputs, enhancing its performance on complex vision-language tasks.

\noindent\textbf{Baselines.}
Since there are currently no methods specifically designed for FGVEdit, we apply the LLM editing methods to the LLM within MLLMs as baselines for evaluation.
\textbf{Fine-tuning} adjusts a pre-trained model’s parameters to incorporate new information. While effective, it is computationally intensive and may unintentionally degrade unrelated behaviors.
\textbf{IKE} \cite{ike} is a paradigm that utilizes contextual learning to modify specific factual knowledge in LLMs without altering their parameters. By providing carefully crafted demonstrations as input prompts, IKE guides the model to update its responses to reflect new or corrected information, avoiding the computational costs of fine-tuning.
\textbf{SERAC} \cite{serac} stores new facts in an additional memory, and employs a scope classifier to determine whether the input is related to the facts stored in memory. By utilizing a counterfactual model to answer questions related to the new facts while preserving the original model, SERAC enables targeted modifications without accessing the underlying model parameters, thereby enhancing computational efficiency.
\textbf{MEND} \cite{mend} leverages an auxiliary neural network to convert fine-tuning gradients into targeted updates, facilitating rapid edits while maintaining performance on unrelated inputs. It serves as a scalable solution for model improvement.

\noindent\textbf{Implementation details.}
For SERAC, MEND, and IKE, we leverage the code provided by the EasyEdit~\cite{wang2023easyedit} toolkit, while fine-tuning is carried out using KEBench~\cite{kebench}. By default, the implementation of our method is built on the SERAC~\cite{serac} framework, utilizing the EasyEdit~\cite{wang2023easyedit} toolkit. To highlight the extensibility of our approach, we replace the original counterfactual model with MEND~\cite{mend} and introduce an enhanced version, called MSCKE-MEND. This demonstrates how easily our method can be adapted to incorporate advanced techniques, resulting in improved performance.

\begin{figure}[t]
  \centering
  \resizebox{\columnwidth}{!}{
   \includegraphics[]{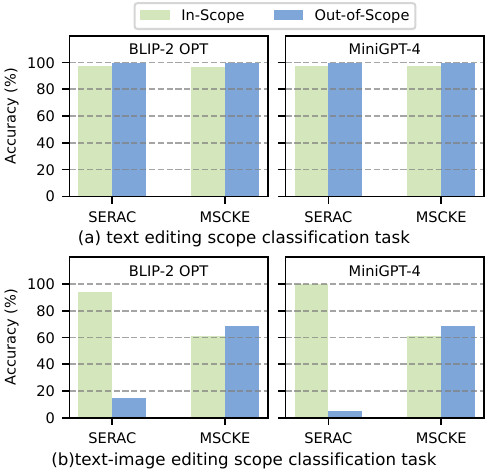}
}
   \caption{Comparison of classification performance between MSCKE's multimodal scope classifier and SERAC's classifier on text and image samples.}
   \label{fig:cls}
\end{figure}

\begin{table}[t]
\centering
\renewcommand{\arraystretch}{1.1}
\resizebox{0.9\columnwidth}{!}{%
\begin{tabular}{lcc}
\hline
Components            & CLIP-ViT-B/32  & CLIP-ViT-L/14  \\ \hline
concatenation         & 63.70          & 63.80          \\
cross-attention       & {\ul 64.45}    & {\ul 64.35}    \\
dot-product attention & \textbf{64.73} & \textbf{64.85} \\ \hline
\end{tabular}%
}
\caption{Implementation analysis of multimodal scope
classifier. Using CLIP-ViT-B/32 for feature extraction, coupled with dot-product attention for feature fusion, yields strong performance.}
\label{tab:ablation}
\end{table}

\begin{table*}[t]
\centering
 \resizebox{0.85\textwidth}{!}{%
\begin{tabular}{llllcccc}
\hline
\multicolumn{2}{c}{source}                             & \multicolumn{2}{c}{target}                             & \multirow{2}{*}{Reliability} & \multirow{2}{*}{Locolity} & \multirow{2}{*}{Generality} & \multirow{2}{*}{Specificity} \\
\multicolumn{1}{c}{method} & \multicolumn{1}{c}{model} & \multicolumn{1}{c}{method} & \multicolumn{1}{c}{model} &                              &                           &                             &                              \\ \hline
MSCKE                    & BLIP-2 OPT                & \multicolumn{1}{c}{/}      & \multicolumn{1}{c}{/}     & 99.13                        & 100                       & 98.56                       & 64.60                        \\
MSCKE                    & BLIP-2 OPT                & MSCKE-MEND               & BLIP-2 OPT                & 97.40/97.40                  & 100/100                   & 96.52/96.50                 & 68.35/68.38                  \\
MSCKE                    & BLIP-2 OPT                & MSCKE                    & MiniGPT-4                 & 99.50/99.50                  & 100/100                   & 93.11/93.00                 & 57.09/57.20                  \\
MSCKE                    & BLIP-2 OPT                & MSCKE-MEND               & MiniGPT-4                 & 97.10/97.05                  & 100/100                   & 96.82/96.70                 & 72.16/71.98                  \\ \hline
\end{tabular}%
 }
\caption{Transferability of the trained multimodal scope classifier to different editing methods and base models. Results are presented as transferred/retrained, where applicable.}
\label{tab:transfer}
\end{table*}

\begin{table}[]
\centering
\tiny
 \resizebox{0.9\columnwidth}{!}{%
\begin{tabular}{lcc}
\hline
               & inference time & model size \\ \hline
classifier     & 36ms           & 0.56G      \\
base           & 121ms          & 9.10G      \\
counterfactual & 85ms           & 4.22G      \\ \hline
\end{tabular}%
 }
\caption{Comparison of inference time and memory usage among the multimodal scope classifier, base model, and counterfactual model on BLIP-2 OPT.}
\label{tab:computational}
\end{table}
\subsection{Results}
Table~\ref{tab:results} shows the results of various methods on the FGVEdit benchmark. We first evaluate the methods across three standard metrics: Reliability, Locality, and Generality. Most editing methods perform excellently in terms of Reliability. However, our methods, MSCKE and MSCKE-MEND, show consistent improvements over their respective baselines, SERAC and MEND, demonstrating superior reliability. For Locality, both FT-LLM and IKE exhibit subpar performance due to the absence of relevant constraints during the editing process. In contrast, only FT-Visual and our MSCKE achieve a perfect score of 100 across both models. This indicates that our method excels in precisely targeting edits without inadvertently altering unrelated information in the text. For Generality, although our method is optimized for multimodal tasks, it outperforms SERAC and MEND even in purely text-based tasks, showcasing the versatility of our approach.

In terms of the specialized metric Specificity, IKE and SERAC show notably limited performance, as they rely exclusively on textual data. Our MSCKE, which incorporates a multimodal classifier leveraging both visual and textual information, significantly outperforms SERAC. Specifically, for BLIP-2 OPT, Specificity increases from 31.92 to 61.60, and for MiniGPT-4, from 37.85 to 57.20. While MEND, which updates both visual and language modules, performs better than other baselines, MSCKE-MEND further enhances performance, improving Specificity from 65.85 to 68.38 for BLIP-2 OPT and from 67.39 to 71.98 for MiniGPT-4. These results underline the effectiveness of our approach in tackling complex multimodal editing tasks.

\subsection{Effect of Multimodal Scope Classifier}
Next, we analyze the performance of our multimodal scope classifier compared to the SERAC scope classifier. Our classifier can process both text-only and text-image inputs, whereas the SERAC classifier is limited to text-only input. We evaluate their editing scope classification performance on two tasks: a text editing scope classification and a text-image editing scope classification task, both using our FGVEdit dataset. The results are shown in Fig.~\ref{fig:cls}. 

For text editing scope classification, both classifiers receive text-only inputs and achieve high accuracy, demonstrating the robustness of our multimodal scope classifier in maintaining strong classification performance for text-based tasks. However, in the text-image editing scope classification, the SERAC scope classifier shows limited effectiveness, largely classifying most samples as in-scope. This limitation arises from the fine-grained nature of our FGVEdit dataset, where both in-scope and out-of-scope samples are closely related to the target editing text. In contrast, our multimodal scope classifier excels by leveraging both textual and visual information, significantly improving classification accuracy.

\subsection{Implementation Analysis of Multimodal Scope Classifier}
The multimodal scope classifier consists of two key components: a feature extraction module and a feature fusion module. We investigate how different implementations of these modules impact the classifier’s performance. The feature extraction module can be implemented using either CLIP-ViT-B/32 or CLIP-ViT-L/14, while the feature fusion module can be realized through feature concatenation, cross-attention, or dot-product attention. The results on FGVEdit dataset are shown in Table~\ref{tab:ablation}.

For feature extraction, CLIP-ViT-B/32 performs similarly to CLIP-ViT-L/14, suggesting that CLIP-ViT-B/32 is sufficient for extracting comprehensive features. This finding highlights that our classifier does not require the more computationally expensive CLIP-ViT-L/14, making the approach more efficient and suitable for practical applications with limited computational resources. Regarding feature fusion, dot-product attention delivers the best performance with minimal computational overhead. Its superior efficiency in capturing essential interactions between text and image features, without the complexity of methods like cross-attention, makes it an ideal choice for feature fusion in our multimodal classifier. 

\subsection{Transferability of Multimodal Scope Classifier}
To further highlight the advantages of our multimodal scope classifier, we investigate its transferability across different editing methods and base models. Specifically, we transfer the multimodal scope classifier, initially trained with FGVEdit dataset on BLIP-2 OPT, to various settings, such as using MEND as the counterfactual model and MiniGPT-4 as the base model. The results are presented in Table~\ref{tab:transfer}. Our multimodal scope classifier demonstrates exceptional transferability. In all experiments, the transferred classifier performs nearly identically to a classifier retrained for each specific setting. This decoupled design and robust transferability enable our method to be quickly adapted for deployment across new MLLMs.

\subsection{Computational Cost Analysis}
The MSCKE framework comprises a multimodal scope classifier, a base model, and a counterfactual model. We evaluate the computational cost of these components in terms of inference time and memory usage. The results on BLIP-2 OPT are shown in Table~\ref{tab:computational}.

Our analysis shows that the multimodal scope classifier requires significantly less inference time and memory compared to both the base model and counterfactual model. These findings highlight that the classifier introduces minimal computational overhead while playing a crucial role in improving editing performance. Its lightweight design ensures that it does not become a bottleneck, making the MSCKE framework well-suited for real-world applications where computational resources may be limited.

%% file: sec/conclusion.tex
\section{Conclusion}
This paper presents a significant advancement in the field of multimodal knowledge editing through the introduction of the Fine-Grained Visual Knowledge Editing~(FGVEdit) benchmark and the Multimodal Scope Classifier-based Knowledge Editor~(MSCKE) framework. By addressing the unique challenges posed by multimodal contexts, our approach enables precise editing while retaining the integrity of unrelated content. The experimental results demonstrate that MSCKE outperforms existing text editing approaches, showcasing its potential to enhance the accuracy and efficiency of knowledge updates in MLLMs. 
